\def\year{2020}\relax
\documentclass[letterpaper]{article} 
\usepackage{aaai18}  
\usepackage{times}  
\usepackage{helvet}  
\usepackage{courier}  
\usepackage{url}  
\usepackage{graphicx}  
\usepackage{algorithm}
\usepackage{amsmath}
\usepackage{algorithmicx}
\usepackage{algpseudocode}
\usepackage{pdfpages}
\usepackage{comment}
\usepackage{caption}
\usepackage{subcaption}
\usepackage{listings}
\usepackage{xcolor}
\begin{document}
%
\title{Stance in Replies and Quotes (SRQ): A New Dataset For Learning Stance in Twitter Conversations}

\author{Ramon Villa-Cox, Sumeet Kumar, Matthew Babcock, Kathleen M. Carley\\
School of Computer Science\\
Carnegie Mellon University\\
5000 Forbes Ave, Pittsburgh, PA 15213, USA\\
}

\maketitle

\begin{abstract}
Automated ways to extract stance (denying vs. supporting opinions) from conversations on social media are essential to advance opinion mining research. Recently, there is a renewed excitement in the field as we see new models attempting to improve the state-of-the-art. However, for training and evaluating the models, the datasets used are often small.  Additionally, these small datasets have uneven class distributions, i.e., only a tiny fraction of the examples in the dataset have favoring or denying stances, and most other examples have no clear stance. Moreover, the existing datasets do not distinguish between the different types of conversations on social media (e.g., replying vs. quoting on Twitter). Because of this, models trained on one event do not generalize to other events. 

In the presented work, we create a new dataset by labeling stance in responses to posts on Twitter (both replies and quotes) on controversial issues. To the best of our knowledge, this is currently the largest human-labeled stance dataset for Twitter conversations with over 5200 stance labels. More importantly, we designed a tweet collection methodology that favours the selection of denial-type responses. This class is expected to be more useful in the identification of rumours and determining antagonistic relationships between users. Moreover, we include many baseline models for learning the stance in conversations and compare the performance of various models. We show that combining data from replies and quotes decreases the accuracy of models indicating that the two modalities behave differently when it comes to stance learning.
\end{abstract}

\section{Introduction}
\begin{figure}[t]
    \centering
    \includegraphics[width=0.50\textwidth]{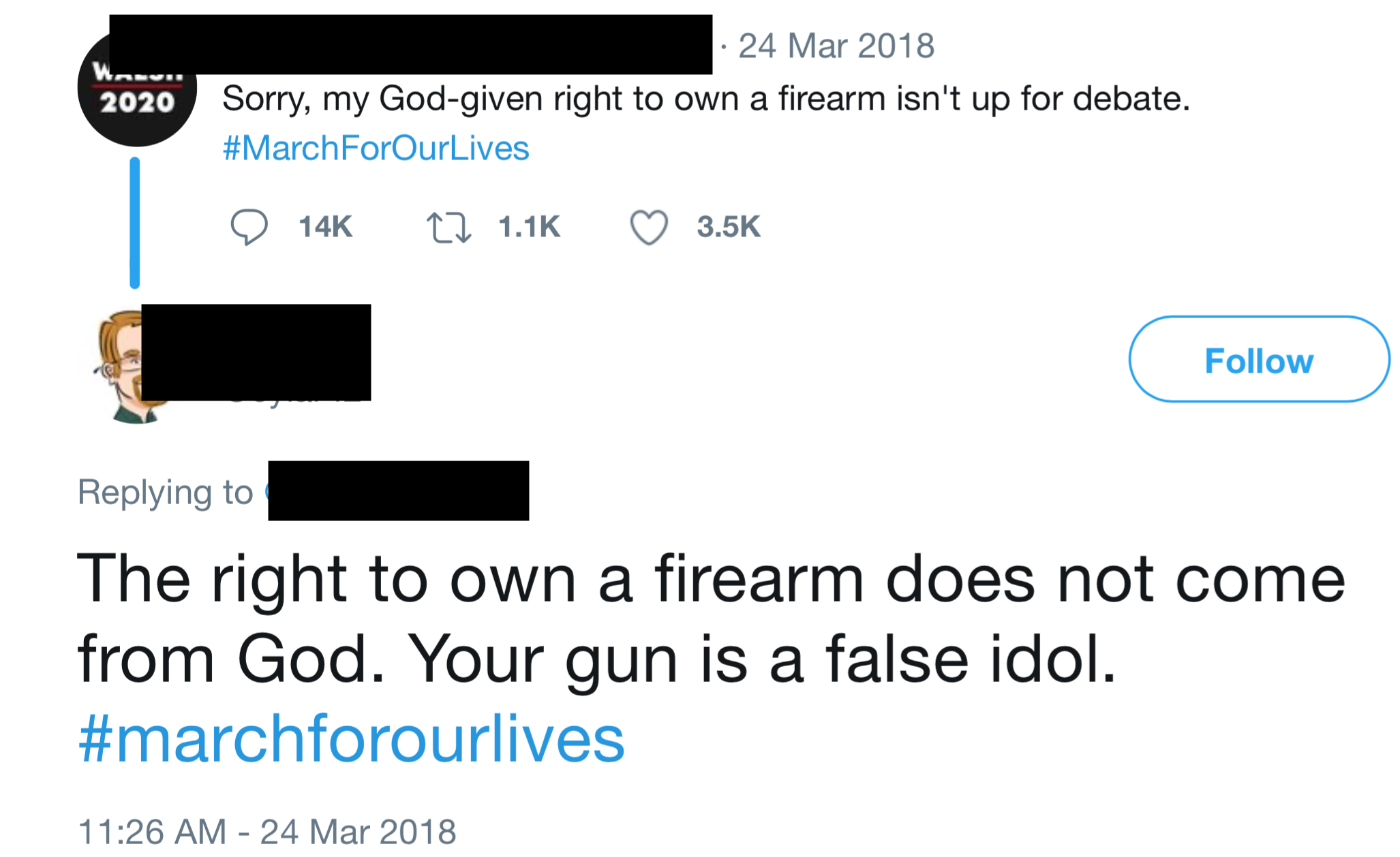}
    \caption{When we reply on Twitter, sometimes we also support or deny others claims. For example, in the conversation shown above, a user denies the claim made in the original tweet. In this research, we build a new dataset to learn the language pattern that users' employ while taking a stance (support vs deny). This dataset could be used to develop automated methods to infer the stance in replies (and quotes).}
    \label{fig:twitter_thread}
\end{figure}

People express their opinions on blogs and other social media platforms. Automated ways to understand the opinions of users in such user-generated corpus are of immense value. It is especially essential to understand the stance of users, which involves finding people's opinions on controversial topics. Therefore, it's not surprising that many researchers have explored automated ways to learn stance given a text \cite{hasan2013stance}. While learning stance from users' individual posts have been explored by several researchers \cite{mohammad2017stance,constance}, there is an increased interest in learning stance from conversations. For example, as we show in Fig. \ref{fig:twitter_thread}, a user denies the claim made in the original tweet. This kind of stance learning has many applications, including insights into conversations on controversial topics \cite{Garimella:2018:QCS:3178568.3140565} and finding potential rumor posts on social-media \cite{zubiaga2015crowdsourcing,zubiaga2018discourse,Babcock2019}. However, the existing datasets used for training and evaluating the stance learning models limit the broader application of stance in conversations.

The existing research on stance in conversations has three significant limitations: 1) The existing datasets are built around rumor events to determine the veracity of a rumor post based on stance taken in replies \cite{zubiaga2015crowdsourcing}. Though useful for rumor detection, this does not generalize to non-rumor events \cite{buntain2017automatically}, 2) The existing datasets focus primarily in direct responses and do not take into account quotes. This is critical as quotes have been gaining prominence since their introduction by Twitter in 2015, especially in the context of political debates \cite{garimella2016quote},  3) The existing datasets have uneven class distributions, i.e., only a small fraction of the examples in the dataset have supporting and denying stances, and most other examples have no clear stance. These unbalanced classes lead to poor learning of denying stance (class) \cite{kumar2019tree}. The denying class is expected to be more useful for downstream tasks like finding an antagonistic relationship between users. Therefore there is a need to build a new dataset that has more denying stance examples.

To overcome the above limitations, in this research, we created a new dataset by labeling the stance in replies (and quotes) to posts on Twitter. To construct this dataset, we developed a new collection methodology that is skewed towards responses that are more likely to have a denial stance. This methodology was applied across three different contentious events that transpired in the United States during 2018. We also collected an additional set of responses without regard to a specific event. We then
labeled a representative sample of the response-target pairs for their stance. Focusing on the identification of denial in responses is an essential step for the identification of tweets that promote misinformation \cite{zubiaga2015crowdsourcing,zubiaga2016analysing} and also to estimate community polarization \cite{Garimella:2018:QCS:3178568.3140565}. By leveraging these human-labeled examples, along with more unlabeled examples on social-media, we expect to build better systems for detecting misinformation and understanding of polarized communities. 


To summarize, the contribution of this work is fourfold: 
\begin{enumerate}
    \item We created a stance dataset (target-response pairs) for three different contentious events (and many additional examples from unknown events). To the best of our knowledge, this is currently the largest human-labeled stance dataset on Twitter conversations with over 5200  stance labels.
    
    \item To the best of our knowledge, this is the first dataset that provides stance labels for Quotes (others are based on replies). This provides a new opportunity to understand the use of quotes.
    
    \item The denial class is the minority label in existing datasets built in a prior research \cite{zubiaga2015crowdsourcing} and is the most difficult to learn, but is also the most useful class for downstream tasks like  rumor detection.  Our method of selecting data for annotation results in a more balanced dataset with a large fraction of support/denial as compared to other stance classes.
    
    \item We introduce two new stance categories by distinguishing between explicit and implicit non-neutral responses. This can help the error analysis of trained classifiers as the implicit class, for either support or denial, is more context dependent and harder to classify.
\end{enumerate}

This paper is organized as follows. We first discuss the related work and then describe our approach to collect the potential tweets to label in `Dataset Collection Methodology'. As the sample that can be labeled is rather small (because of budget limitations) compared to the entire available dataset, we discuss the sample construction procedure for annotation. Then, we describe the annotation process and the statistics of the dataset that obtained as a result of annotation in section `Annotation Procedure and Statistics'. Next, we present some baseline models for stance learning  and present the result. Finally, we discuss our results and propose future directions.





\section{Related Work}
\label{sec:RelatedWork}
Topics on learning stance from data could be broadly categorized as having to do with: 1) Stance in posts on social media, and 2) Stance in Online Debates and Conversations. We next describe prior work on these topics.

\subsection{Stance in Social-Media Posts}
Mohammad et al. \cite{mohammad2017stance} built a stance dataset using Tweets of several different topics, and organized a SemEval competition in 2016 (Task \#6). Many researchers \cite{augenstein2016usfd,liu2016iucl,wei2016pkudblab} used this dataset and proposed algorithms to learn stance from data. However none of them exceeded the performance achieved by a simple algorithm \cite{mohammad2017stance} that uses word and character n-grams, sentiment, parts-of-speech (POS) and word embeddings as features. The authors used an SVM classifier to achieve 0.59 as the mean f1-macro score. While learning stance from posts is useful, the focus of this research is stance in conversations. Conversations allow a different way to express stance on social media in which a user supports or denies a post made by another user. Stance in a post is about authors' stance on any topic of interest (pro/con), in contrast, stance in conversation is about stance taken when interacting (replying or quoting) with other authors (favor/deny). We describe this in detail in the next section.

\subsection{Stance in Online Debates and Conversations}
The idea of stance in conversations is very general and its research origin can be traced back to identifying stance in online debates \cite{somasundaran2010recognizing}.  Stance in online debates have been explored by may researchers recently \cite{sridhar2014collective,hasan2013stance,sobhani2015argumentation}. Though stance-taking by users on social-media, especially on controversial topics, often mimic a debate, social-media posts are very short. An approach of stance mining that combines machine-learning to predict stance in replies  -- categorized as `supporting',  `denying', `commenting' and `querying' -- to a social media post is gaining popularity \cite{zubiaga2016stance,zubiaga2015crowdsourcing}. Prior work has confirmed that replies to a `false' (misleading) rumor are likely to have replies that deny the claim made in the source post \cite{zubiaga2016analysing}. Therefore, this approach is promising for misinformation identification \cite{Babcock2019}. However, the earlier stance dataset on conversations was collected around rumor posts \cite{zubiaga2015crowdsourcing}, and contains only replies, and has relatively few denials. Our new dataset generalizes this approach and extends it to quotes-based  interactions on controversial topics. As described, this new dataset is distinct as: 1) it distinguishes between `replies' and `quotes', the two very different types of interaction on Twitter, 2) it is collected in way to get more `denial' stance examples, which was a minority label in \cite{zubiaga2016stance}, and 3) it is collected on general controversial topics and not on rumor posts.

\section{Dataset Collection Methodology}\label{methodology}
\begin{figure}[ht] 
    \centering
    \includegraphics[width=0.48\textwidth]{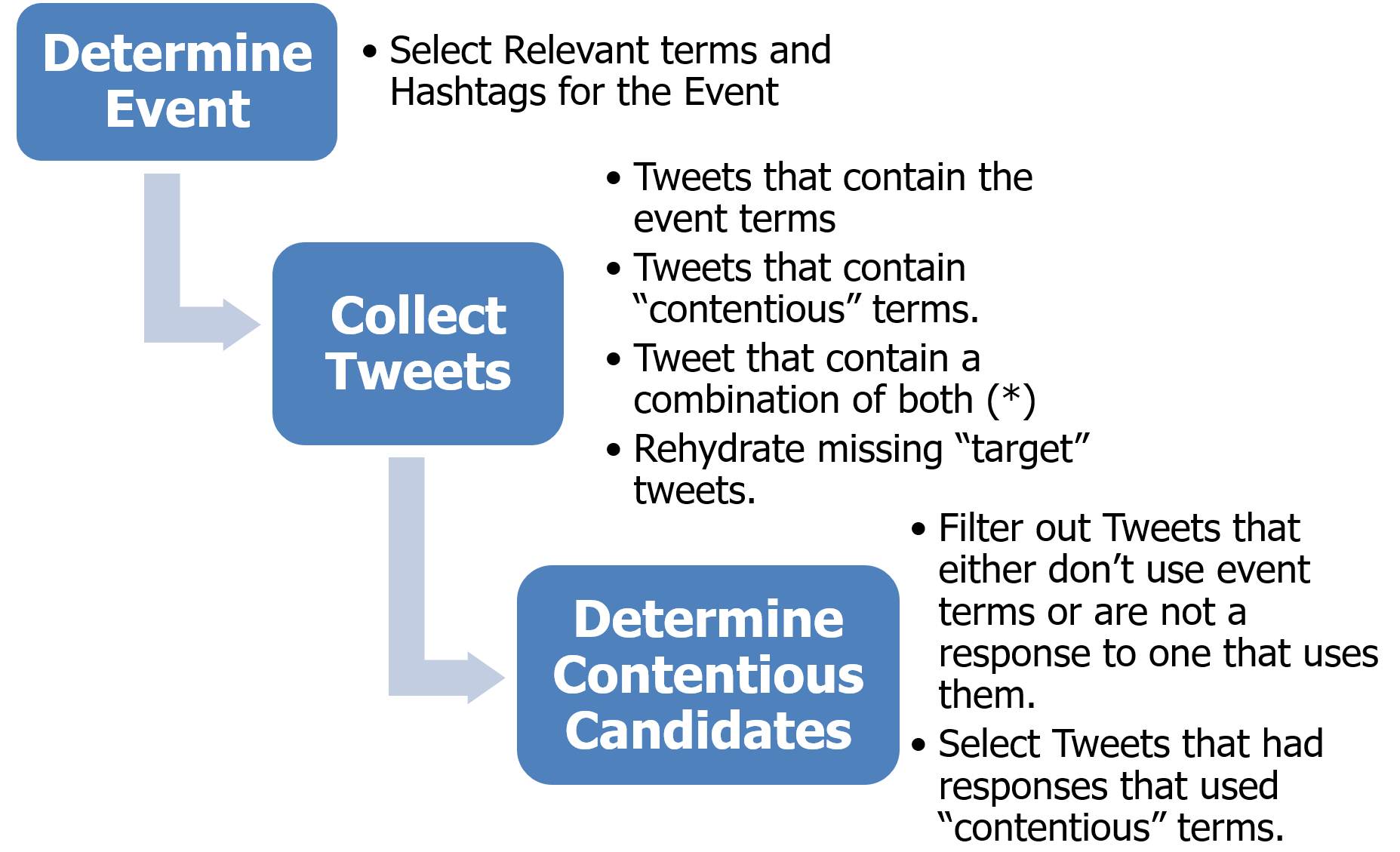}
     \caption{Methodology developed for the collection of contentious tweet candidates for a specific event.}
     \label{fig:col_method}
 \end{figure}

 Figure \ref{fig:col_method} summarizes the methodology developed to construct the datasets that skews towards more contentious conversation threads. We describe the steps in details next.
 
The first step requires finding the event related terms that could be used to collect the source (also called target) tweets. Additionally, as the focus is on getting more replies that are denying the source tweet, we use a set of contentious terms used to filter the responses made to the source tweets.

\subsection{Step 1: Determine Event}
The collection process centered on the following events.
\begin{itemize}
    \item \textbf{Student Marches}:  This event is based on the `March for Our Lives' student marches that occurred on the 24 of March of 2018 in the United States. Tweets were collected from March 24 to April 11 of 2018. \\The following terms were used as search queries: \#MarchForOurLives, \#GunControl, Gun Control, \#NRA, NRA, Second Amendment, \#SecondAmendment.
    
    \item \textbf{Iran Deal}: This event involves the prelude and aftermath of the United States announcement of its withdrawal from the Joint Comprehensive Plan of Action (JCPOA), also known as the "Iran nuclear deal" on May 8, 2018. Tweets were collected from April 15 to May 18 of 2018. \\
    The following terms were used as search queries: Iran, \#Iran, \#IranDeal, \#IranNuclearDeal, \#IranianNuclearDeal, \#CancelIranDeal, \#EndIranNuclearDeal, \#EndIranDeal.
    
    \item \textbf{Santa Fe Shooting}: This event involves the prelude and aftermath of the Santa Fe School shooting that took place in Santa Fe, Texas, USA in May 18, 2018. \\
    Tweets were collected from May 18 to May 29 of 2018. For this event, the following terms were used as search queries: Gun Control, \#GunControl, Second Amendment, \#SecondAmendment, NRA, \#NRA, School Shooting, Santa Fe shooting, Texas school shooting.
    
    \item \textbf{General Terms}: This defines a set of  tweets collected that were not from any specific event, but are collected based on responses that contain the contentious terms described next. Tweets were collected from July 15 to July 30 of 2018.
    
\end{itemize}

The set of contentious terms used across all events are divided in 3 groups: hashtags, terms and fact-checking domains:
\begin{itemize}
    \item \textbf{Hashtags}: \#FakeNews, \#gaslight, \#bogus, \#fakeclaim, \#deception, \#hoax, \#disinformation, \#gaslighting.
    \item \textbf{Terms}: FakeNews, bull**t, bs, false, lying, fake, there is no, lie, lies, wrong, there are no, untruthful, fallacious, disinformation, made up, unfounded, insincere, doesnt exist, misrepresenting, misrepresent, unverified, not true, debunked, deceiving, deceitful, unreliable, misinformed, doesn't exist, liar, unmasked, fabricated, inaccurate, gaslight, incorrect, misleading, deception, bogus,  gaslighting, mistaken, mislead, phony, hoax, fiction, not exist.
    \item \textbf{URLs}: www.politifact.com, www.factcheck.org, www.opensecrets.org, www.snopes.com.
\end{itemize}

\subsection{Step 2: Collect Tweets}
Using Twitter's REST and the Streaming API we collected tweets that used either the event or contentious terms (as described earlier). If the target of a response is not included in the collection, we obtained it from Twitter using their API. 


\subsection{Step 3: Determine Contentious Candidates}
A target-response pair is selected as potential candidate to label if the target contains any of the listed event terms and the response contains any of the contentious terms. If urls are in the tweet, they are matched at the domain level by using the \textit{urllib} library in Python. For  `General Terms' event collected pairs based solely on the responses regardless of the terms used in the target.

To reduce the sample size, we filtered the tweets on some additional conditions. We only used the responses that were identified by Twitter to be in English and excluded responses from a user to herself (as this are used to form threads). In order to simplify the labeling context, we also excluded responses that included videos, or that had targets that included videos and limited our sample set to responses to original tweets. This effectively limits the dataset to the first level of the conversation tree.

The above steps resulted in a dataset which can potentially be labeled. We show the distribution of this dataset in Tab. \ref{tbl:universe}. Because this set is large, we developed a method to a retrieve a smaller sample for labeling. We describe this sample construction method next.

\begin{center}
\begin{table}[htb]
 \begin{tabular}{|p{2.7cm}|p{1.5cm}|p{1.5cm}|} 
 \hline
 Event  & Replies & Quotes \\ [0.5ex] 
 \hline
 Student Marches   & 23314 & 8321 \\
 \hline
 Santa Fe Shooting   & 24494 & 11825\\
 \hline
 Iran Deal   & 21290  & 14939 \\
 \hline
 General Terms  & 3756269  & 2540084 \\
  \hline
\end{tabular}
\caption{Distribution of relevant tweet pairs by response type that could be labeled.}
\label{tbl:universe}
\end{table}
\end{center}


 


 
\section{Sample Construction for Annotation}\label{sec:sample}
We sought to design a sample that was representative of the semantic space observed on the responses across the different events. For this purpose we encoded the collected responses via Skip-Thought vectors \cite{kiros2015skip}, to obtain an a priori semantic representation. The Skip-Thought model is trained using a large text dataset such that the vector representation of the text encodes the meaning of the sentence. To generate vectors, we use the pre-trained model shared by the authors of Skipthought \footnote{https://github.com/ryankiros/skip-thoughts}. The model uses a neural-network that takes text as input and generate a 4800 dimension embedding vector for each sentence. Thus, on our dataset, for each response in Twitter conversations, we get a 4800 dimension vector representing the semantic space.  

\begin{figure}[htb!] 
    \centering
    \includegraphics[width=0.46\textwidth]{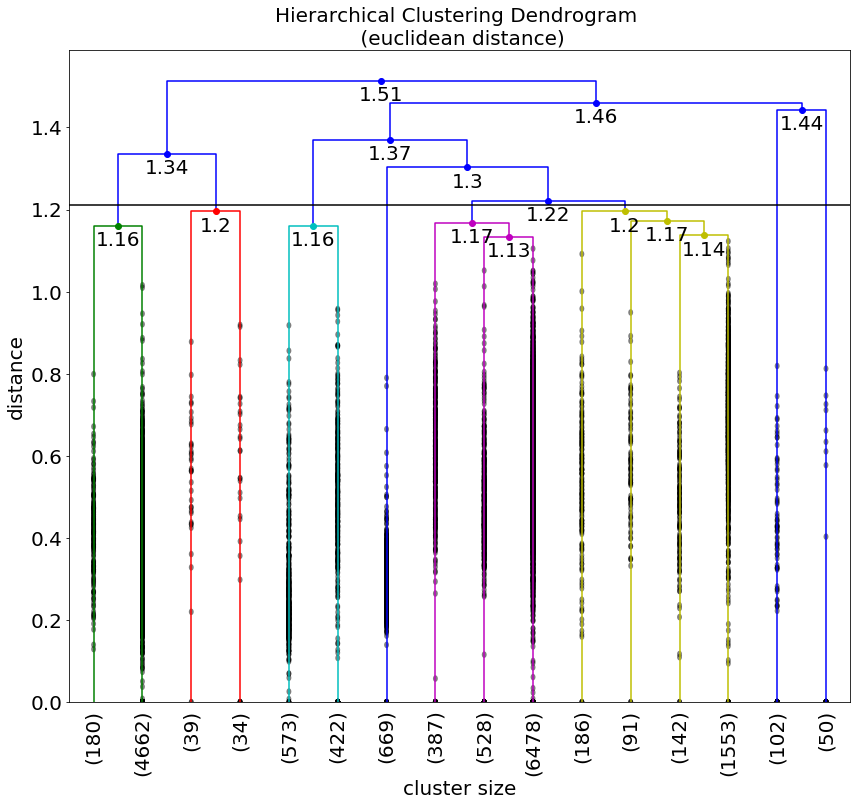}
     \caption{Dendogram derived for the Student Marches event. Horizontal line describes the maximum cophenetic distance used when determining the final cluster labels. Further bifurcations of the dendogram where replaced with dots in order to avoid clutter.}
     \label{fig:dendogram}
 \end{figure}
 
To obtain a representative sample of the semantic space, we applied a stratified sampling methodology \footnote{Stratified Sampling is a sampling method that divides a population in exhaustive and mutually exclusive groups which can reduce the variance of estimated statistics.}. The strata were determined by clustering the space via hierarchical clustering methods using a 'average' linkage algorithm and a euclidean distance metric. It is important to note that given the difficulty of assessing clustering quality on such high-dimensional spaces (over 4k dimensions), we first reduced the space to 100 dimensions via Truncated Stochastic Value Decomposition \cite{xu1998truncated}.  Figure \ref{fig:dendogram} presents the derived dendogram and the optimal number of clusters selected for the Student Marches event, a similar analysis was done for the other events. The relevant hyper-parameters used were determined by evaluating the final clustering quality based on the resulting cophenetic correlation \cite{saraccli2013comparison}. It is important to note that the number of clusters selected was higher than the optimal, as our main purpose is to get a thorough partition of the semantic space.

 A two level stratified scheme was utilized, with the second level being the type of response. This means that the percentage of Quotes and Replies within each stratum were maintained. Finally, we decided to under-sample, by a factor of two, the responses to verified accounts so that our final sample has more interaction between regular Twitter users. The final sample distribution by response type is presented in table \ref{tbl:sample}.
 
 \begin{center}
\begin{table}[htb]
 \begin{tabular}{|p{3.7cm}|p{1.2cm}|p{1.2cm}|} 
 \hline
 Event  & Replies & Quotes \\ [0.5ex] 
 \hline
 Student Marches (SM)  & 293 & 443 \\
 \hline
 Santa Fe Shooting  (SS) & 609 & 609\\
 \hline
 Iran Deal  (ID) & 508  & 738 \\
 \hline
 General Terms  (GT) & 1476  & 544 \\
  \hline
\end{tabular}
\caption{Distribution of relevant tweet pairs by response type. Notice that these terms tend to be used more frequently in direct replies.}
\label{tbl:sample}
\end{table}
\end{center}

\begin{figure*}[ht] 
    \centering
    \includegraphics[width=0.45\textwidth]{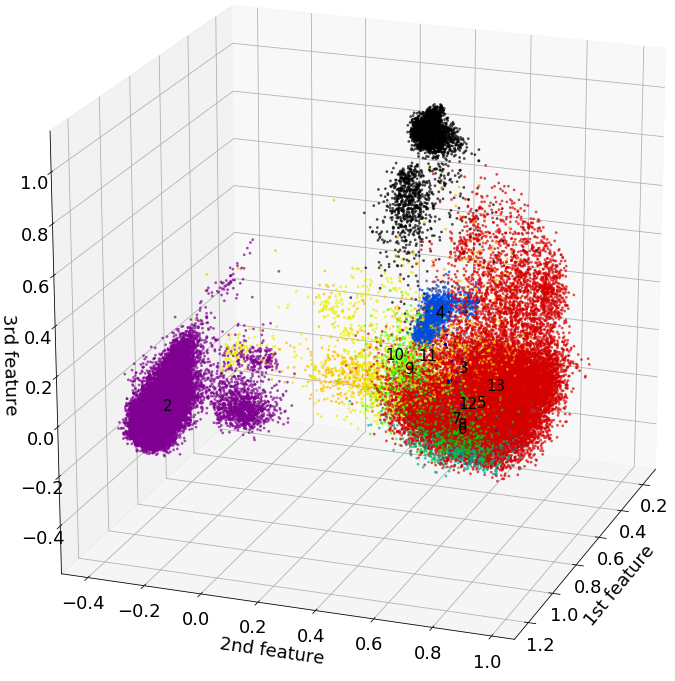}
    \includegraphics[width=0.45\textwidth]{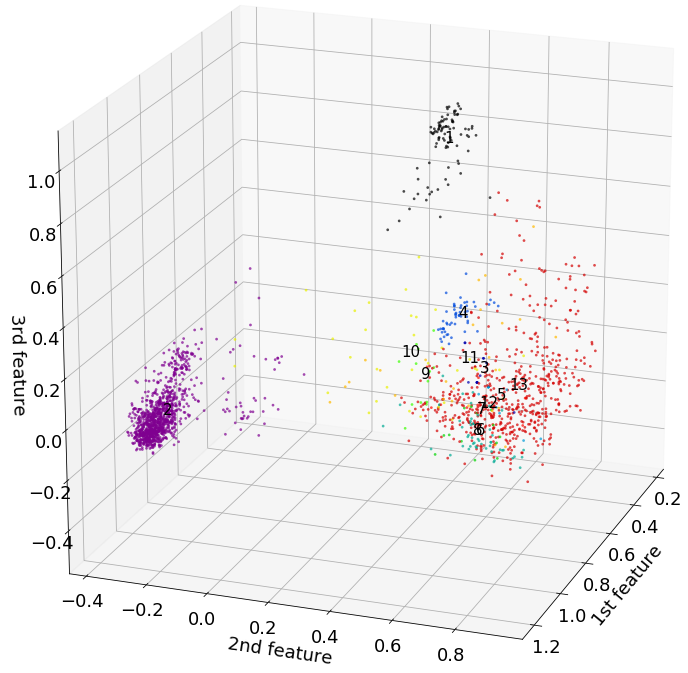}
     \caption{3-dimensional representation, obtained via Truncated Stochastic Value Decomposition,  of the skip-thought vector representation for the responses in the General Terms event. The top figure corresponds to the collected universe and the bottom to the derived sample. Similar distributions and clustering behavior is observed on other events.}
     \label{fig:3_dim}
 \end{figure*}

Figure \ref{fig:3_dim} presents a 3-dimensional representation, obtained via Truncated Stochastic Value Decomposition,  of the semantic space observed for the responses in the General Terms event and the derived sample. A similar clustering pattern is observed on other events as well. Notice that the sample covers fairly well the observed semantic distribution, especially when compared with simple random sampling.

\section{Annotation Procedure and Statistics}
Recent work on stance labeling in social media conversations has centered on identifying 4 different positions in responses: agreement, denial, comment, and queries for extra information \cite{procter2013reading,zubiaga2016analysing}. We introduced two extra categories, by distinguishing between explicit and implicit  non-neutral responses. The former refers to responses that include terms that explicitly state that their target is wrong\textbackslash right (e.g. `That is a blatant lie!'). The implicit category on the other hand, as its name implies, correspond to responses that do not explicitly mention the stance of the user, but that, given the context of the target, are understood as denials or agreements. These are much harder to classify, as they can include sarcastic responses.

\begin{figure*}[ht!]
    \centering
    \frame{\includegraphics[width=0.9\textwidth]{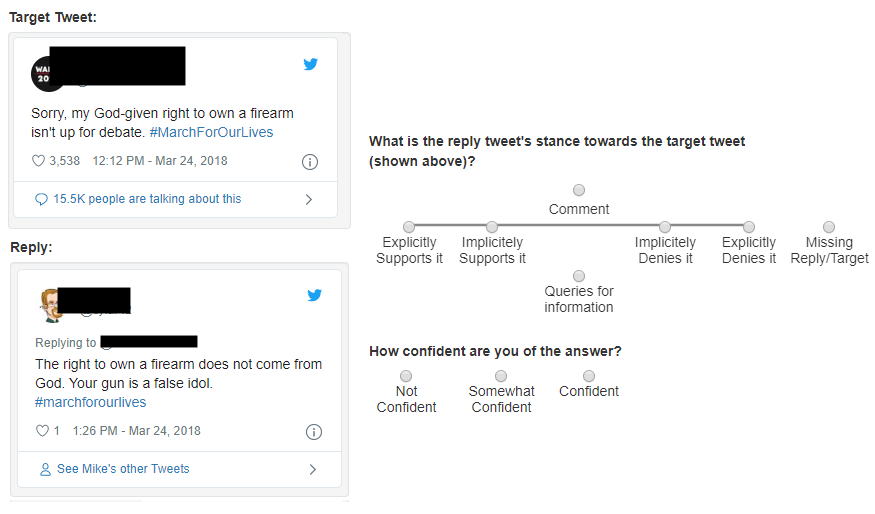}}
     \caption{Snapshot of the webpage developed for annotating replies. Annotators are required to provide the stance in the reply and their confidence in the provided label. }
      \label{fig:annex_pages}
\end{figure*}

The annotation process was handled internally by our group and for this purpose we developed a web interface for each type of response (see Fig. \ref{fig:annex_pages}). Each annotator was asked to go through a tutorial and a qualification test to participate in the the annotation exercise. The annotator is required to indicate the stance of the response (one of the six options in the list below) towards the target and also provide a level of confidence in the label provided. If the annotator was not confident in the label, then the task was passed to another annotator. If both labels agreed, the label was accepted and if not the task was passed to a third annotator. Then the majority label was assigned to the response, and in the few cases were disagreement persisted, the process was continued with a different annotator until a majority label was found.

\subsection{Definition of Classes}
We define the stance classes as:
\begin{enumerate}
    \item Explicit Denial: Explicitly Denies means that the quote/tweet outright states that what the target tweets says is false.
    \item Implicit Denial: Implicitly Denies means that the quote/tweet implies that the tweeter believes that what the target tweet says is false.
    \item Implicitly Support: Implicitly Supports means that the quote/tweet implies that the tweeter believes that what the target tweet says is true.
    \item Explicitly Support: Explicitly Supports means that the quote/tweet outright states that what the target tweets says is true.
    \item Queries: Indicates if the reply asks for additional information regarding the content presented in the target tweet.
    \item Comment: Indicates if the reply is neutral  regarding the content presented in the target tweet.
\end{enumerate}

\begin{figure}[ht!]
    \centering
    \includegraphics[width=0.45\textwidth]{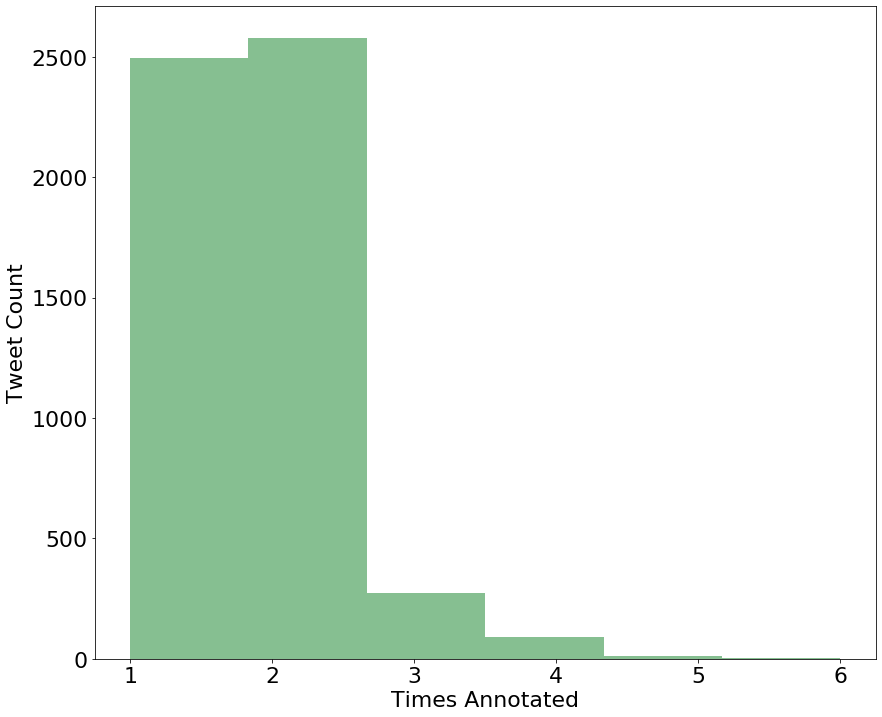}
     \caption{Histogram of number of times tweets were annotated. As we used confidence score in labeling, the labeled tweets for which the labler had low confidence were relabeled by more labelers. This process resulted in some tweets getting labeled up to 5 times to obtain confidence in the assigned class label.}
      \label{fig:annex_counts}
\end{figure}

To validate the methodology, we selected 55\% of the tweets that were initially confidently labeled to be annotated again by a different team member. Of this sample, 86.83\% of the tweets matched the original label and the remainder required additional annotation to find a majority consensus. From the 13.17\% of inconsistent tweets, a 61.86\% were labeled confidently by the second annotator. This means that among the confident labels we validated, only 8.15\% resulted in inconsistencies between two confident annotators, which we deemed an acceptable error margin. Figure \ref{fig:annex_counts}, shows the distribution of times the tweets were annotated. As shown, 45\% of tweets were annotated only once, 47\% were annotated twice, 5\% were annotated three times and less than 2\% required more than three annotations.

Figure \ref{fig:annex_counts}, shows the distribution of times the tweets were annotated. As shown, 45\% of tweets were annotated only once, 47\% were annotated twice, 5\% were annotated three times and less than 2\% required more than three annotations.

 \begin{center}
\begin{table}[htb]
 \begin{tabular}{|p{2.3cm}|p{1.1cm}|p{0.8cm}|p{1.3cm}|p{1.1cm}|} 
 \hline
   & General Terms & Iran Deal & Santa Fe Shooting & Student Marches\\ [0.5ex] 
\hline
Comment & 656 & 293 & 246 & 153 \\
\hline
Explicit Denial & 521 & 350 & 471 & 253 \\
\hline
Implicit Denial & 202 & 116 & 116 & 49 \\
\hline
Explicit Support & 138 & 118 & 85 & 47 \\
\hline
Implicit Support & 415 & 327 & 279 & 215 \\
\hline
Queries & 88 & 42 & 21 & 19 \\
\hline
\end{tabular}
\caption{Distribution of labels across different events.}
\label{tbl:labels}
\end{table}
\end{center} 

Table \ref{tbl:labels} presents the label distribution for the different events. As expected we observe that the labeled dataset is skewed towards denials as, when combining implicit and explicit types, they constitute the majority label for all events. Interestingly, when applied to a specific event, the "comment" category fall behind the two explicit non-neutral labels. This suggest that for contentious events, the proposed collection methodology is effective at recovering contentious conversations and more non-neutral threads. 

\begin{figure}[htb!] 
    \centering
    \includegraphics[width=0.48\textwidth]{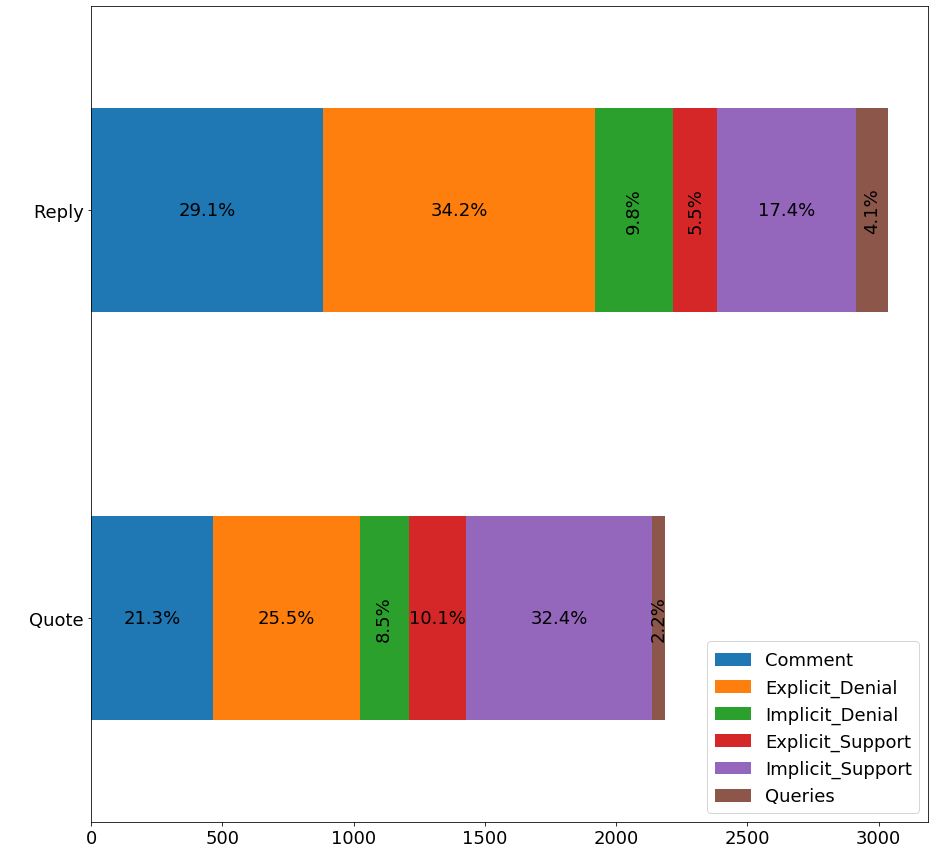}
     \caption{Distribution of the labels among the different response types. Note that among Quotes, the majority label is implicit support. This shows how these type of responses tend to be more context dependent and harder to label.}
     \label{fig:labels}
 \end{figure}

In Figure \ref{fig:labels}, we show the distribution of the labels for each type of response. Note that among Quotes, the majority label becomes implicit support, which shows how these types of responses are more context dependent. As we show in the next section, this also translates on a more complex prediction task.

\subsection{Inter Annotator Agreement}

To validate the methodology, we selected 55\% of the tweets that were initially confidently labeled to be annotated again by a different team member. Of this sample, 86.83\% of the tweets matched the original label and the remainder required additional annotation to find a majority consensus. From the 13.17\% of inconsistent tweets, a 61.86\% were labeled confidently by the second annotator. This means that among the confident labels we validated, only 8.15\% resulted in inconsistencies between two confident annotators, which we deemed an acceptable error margin.

Cohen’s kappa measures the agreement between two or more raters. If each rate labels N items into C categories, Cohen kappa is defined as:

\begin{equation}
\kappa = \frac{p_0 - p_e}{ 1 - p_e}     \\
= \frac{0.92 - 0.33}{ 1 - 0.33} \\
= 0.89
\end{equation}

where $p_0$ is the relative observed agreement among raters and $p_e$ is the estimate of possible agreement by chance. In our experiment, $p_0 = 0.92$ and agreement chance $p_e = 0.33$ as there are three class types. This leads to $\kappa$ value of 0.89.

\subsection{Distribution of Users' Stance}
In addition to conversations, we also labeled a small set of users in the dataset for their stance. For `Santa Fe Shooting' and `Student Marches', the stance was labeled for `Pro/Con' gun control. For `Iran Deal', the stance was evaluated for pro and against the breaking of the Iran deal agreement. The labeled dataset is summarized in Tbl. \ref{tbl:labels_users}.

\begin{center}
	\begin{table}[htb]
		\caption[Distribution of labeled users' stance]{Distribution of labeled users' stance.}
		\centering
		\begin{tabular}{|p{1.3cm}|p{1.1cm}|p{1.3cm}|p{1.1cm}|} 
			\hline
			& Iran Deal & Santa Fe Shooting & Student Marches\\ [0.5ex] 
			\hline
			\hline
			Pro &  137 & 188 &129 \\
			\hline
			Anti & 122  & 64 & 154\\
			\hline
		\end{tabular}
		
		\label{tbl:labels_users}
	\end{table}
\end{center}

\section{Dataset Schema and FAIR principles}
In adherence to the FAIR principles, the database was uploaded to Zenodo and is accessible with the following link \url{http://doi.org/10.5281/zenodo.3609277}. We also adhere to Twitter's terms and conditions by not providing the full tweet JSON but provide the tweet ID so that it can be rehydrated. However, for the labeled tweets, we do provide the text of the tweets and other relevant metadata for the reproduction of the results. The annotated tweets are included in a JSON file with the following fields:
\begin{itemize}
    \item \textit{event}: Event to which the target-response pair corresponds to.
    \item \textit{response\_id}: Tweet ID of the response, which also served as the unique and eternally persistent identifier of the labeled database (in adherence to principle F1).
    \item \textit{target\_id}: Tweet ID of the target.
    \item \textit{interaction\_type}: Type of Response: Reply or Quote.   
    \item \textit{response\_text}: Text of the response tweet.
    \item \textit{target\_text}: Text of the target tweet.
    \item \textit{response\_created\_at}: Timestamp of the creation of the response tweet.
    \item \textit{target\_created\_at}: Timestamp of the creation of the target tweet.
    \item \textit{Stance}: Annotated Stance of the response tweet. The annotated categories are: Explicit Support, Implicit Support, Comment, Implicit Denial, Explicit Denial and Queries. 
    \item \textit{Times\_Labeled}: Number of times the target-response pair was annotated. 
\end{itemize}

We also include a separate dataset that provides the universe of tweets from which the labeled dataset was selected. Because of the number of tweets involved, we do not include the text of the target-response pairs. These tweets are included in a JSON file with the following fields:
\begin{itemize}
    \item \textit{event}: Event to which the target-response pair corresponds to.
    \item \textit{response\_id}: Tweet ID of the response.
    \item \textit{target\_id}: Tweet ID of the target.
    \item \textit{interaction\_type}: Type of Response: Reply or Quote.       
    \item \textit{response\_text}: Text of the response tweet.
    \item \textit{terms\_matched}: List of 'contentious' terms found on the text of the response tweet. 
\end{itemize}

 
 
\section{Baseline Models and Their Performance}
We consider a number of classifiers including traditional text features based classifiers and neural-networks (or deep learning) based models. In this section, we describe the input features, the model architecture details, the training process and finally, discuss the results.

\subsection{Input Features}
As we have sentence pairs as input, we use features extracted from text to train the models. For each sentence pair, we extract text features from both the source and the response separately. 


\subsubsection{TF-IDF}
Tf-Idf (Term frequency- inverse document frequency) \cite{salton1988term} is very popular feature commonly used in many text based classifier. In our research, we use TF-IDF along with Support-Vector Machine (SVM) model that we describe later.

\subsubsection{Glove (GLV)}
 In this kind of sentence encoding, word vectors are obtained for each word of a sentence, and the mean of these vectors are used as the sentence embedding.  To get word vectors, we used Glove \cite{pennington2014glove} which is one the most commonly used word vectors. Before extracting the Glove word vectors, we perform some basic text cleaning which involves removing any @mentions, any URLs and the Twitter artifact (like `RT') which gets added before a re-tweet. Some tweets, after cleaning did not contain any text (e.g. a tweet that only contains a URL or an @mention). For such tweets, we generate an embedding vector that is an average of all sentence vectors of that type in the dataset. The same text cleaning step was performed before generating features for all embeddings described in the paper.
 
\subsubsection{Skip-thoughts (SKP)}
We use the pre-trained model shared by the authors of Skipthought \footnote{https://github.com/ryankiros/skip-thoughts}. The model uses a neural-network that takes sentences as input and generate a 4800 dimension embedding for each sentence \cite{kiros2015skip}. Thus, on our dataset, for each post in Twitter conversations, we get a 4800 dimension vector

\subsubsection{DeepMoji (DMJ)}
We use the DeepMoji pre-trained model \footnote{https://github.com/huggingface/torchMoji} to generate deepmoji vectors \cite{felbo2017using}. Like skipthought, DeepMoji is a neural network model that takes sentences as input and outputs a 64 dimension feature vectors. 

The process of training the LSTM model using DeepMoji vectors closely follows the training process  for the semantic features. The only difference is that the input uses DeepMoji vectors, and hence the size of input vector changes.

\subsection{Classifiers}
As mentioned earlier, we tried two types of classifiers: 1) TF-IDF Text features based classifiers, and 2) neural-networks (deep learning) based classifiers. For the classification task, we only consider four class classification by merging `Explicit Denial' and `Implicit Denial' as Denial, and `Implicit Support' and `Explicit Support' as Support. We describe the details of the classifiers next.

\subsubsection{SVM with TF-IDF features}
Support Vector Machine (SVM) is  a classifier of choice for many text classification tasks. The classifier is fast to train and performs reasonably well on wide-range of tasks. For the Text SVM classification, we only use the reply text to train the model. The classifier takes TF-IDF features as input and predicts the four class stance classes. We would expect that this simple model cannot effectively learn to compare the source and the reply text as is needed for good stance classification. However, we find that such models are still very competitive and therefore serves as a good baseline. 

\subsubsection{Deep Learning models with GLV, SKP, DMJ features}

\begin{figure}[htb!]
    \centering
    \includegraphics[width=0.42\textwidth]{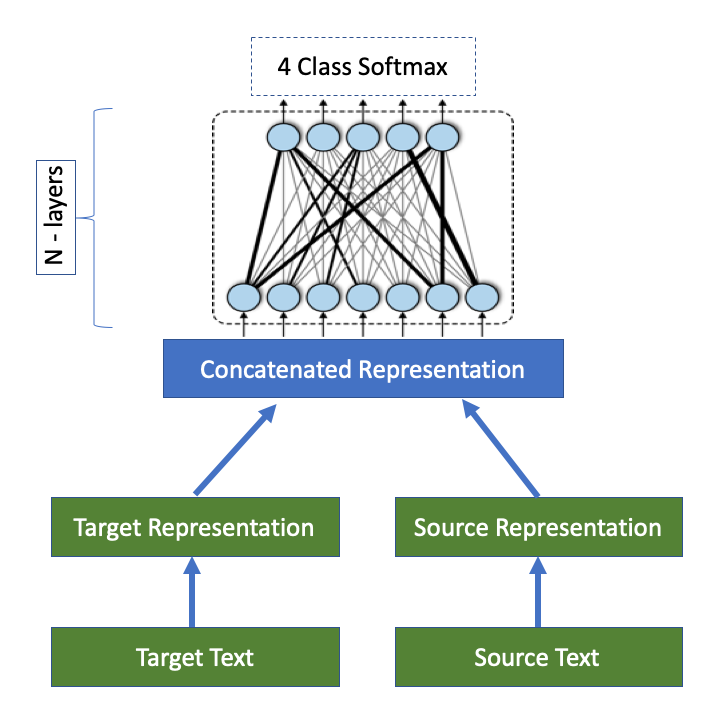}
    \caption{Deep learning model sample diagram}
    \label{fig:model_digram}
\end{figure}

As opposed to traditional text classifiers, neural-network based models could be designed to effectively use text-reply pair as input. One such model is shown in Fig. \ref{fig:model_digram}. A neural network based architecture that uses both source and reply can effectively compare target and reply posts and we  expect it to result in a better performance. This type of neural network can further be divided in two types based on inputs: 1) Word vectors (or embeddings) are used as input such as Glove (GLV), 2) Sentence vectors (or sentence representations) are used as input such as skip-thoughts, DeepMoji and a joint representation of skip-thought and deep-moji (SKPDMJ).  The first model that takes word embeddings as input requires a recurrent layer that embeds the text and reply to a fixed vector representation (one for target and one for reply). One fully connected layer that uses the fixed vector representation input and a softmax layer on top to predict the final stance label. The second type of model that uses the text and reply representations only have one (or more) fully connected layer and a softmax layer on top to predict the final stance label.

\begin{center}
\begin{table*}[htb!]
\small
 \begin{tabular}{|p{2.2cm}| 
 p{0.55cm}|p{0.55cm}|p{0.55cm}|
 p{0.55cm}|p{0.55cm}|p{0.55cm}|
 p{0.55cm}|p{0.55cm}|p{0.55cm}|
 p{0.55cm}|p{0.55cm}|p{0.55cm}|
 p{0.5cm}|p{0.5cm}|p{0.5cm}|} 

 \hline
 {\textit{Model$\downarrow$ Event $\rightarrow$ }} & \multicolumn{3}{|l|}{{\textit{Iran Deal  (ID)}}} &  \multicolumn{3}{|l|}{{\textit{General Terms  (GT)}}} & \multicolumn{3}{|l|}{{\textit{Student Marches (SM)}}}  &
 \multicolumn{3}{|l|}{{\textit{Santa Fe Shooting  (SS)}}} &
 \multicolumn{3}{|l|}{{\textit{Mean}}}\\
 \hline
 Data Type&QOT & RPL &CMB &
 QOT & RPL &CMB &
 QOT & RPL &CMB &
 QOT & RPL &CMB &
 QOT & RPL &CMB \\
 \hline
 \hline
 \multicolumn{16}{|l|}{\textbf{\textit{Baseline Models}}} \\
 \hline
 Majority & 0.46& 0.47 & 0.37 & 0.37& 0.36 & 0.36  & 0.53& 0.50 & 0.41  &  0.40& 0.56 & 0.48 & 0.44& 0.47 & 0.41 \\
 Text SVM & 0.44& 0.44 & 0.43 & 0.46& 0.41 & 0.41  & 0.45& 0.51 & 0.45  &  0.44& 0.55 & 0.48 & 0.45& 0.48 & 0.44 \\
 \hline
 \hline
 \multicolumn{16}{|l|}{\textbf{\textit{ Deep Learning Models}}} \\ \hline
 Glove & 0.41& 0.46 & 0.40 & 0.42& 0.41 & 0.42  & 0.49& 0.48 & 0.47  &  0.47& 0.56 & 0.49 & 0.45& 0.48 & 0.45 \\
 SKP & 0.46& 0.42 & 0.39 & 0.38& 0.37 & 0.37  & 0.48& 0.50 & 0.42  &  0.38& 0.53 & 0.46 & 0.43& 0.45 & 0.41 \\
 DMJ & 0.46& 0.46 & 0.40 & 0.40& 0.39 & 0.41  & 0.54& 0.51 & 0.44  &  0.41& 0.56 & 0.48 & 0.45& 0.48 & 0.43 \\
 SKPDMJ & 0.45& 0.41 & 0.39 & 0.39& 0.39 & 0.36  & 0.46& 0.49 & 0.42  &  0.46& 0.51 & 0.44 & 0.44& 0.45 & 0.40 \\
 \hline
\end{tabular}
\caption{Classification results for Replies: F1-score (micro) and mean of F1 scores (Mean) for different events. QOT implies quotes, RLP implies replies and CMB implies combined quotes and replies.}
 \label{tbl:classification_results_for_all}
\end{table*}
\end{center} 



\subsection{Classifiers Training}
Our neural-network based models are built using Keras library \footnote{https://keras.io/}. The models used feature vectors (Glove, SKP, DMJ) as input. Because Glove is a word vector embeddings, we use a 
recurrent layer right above the input to create a fixed size sentence embeddings vector. For SKP, DMJ and SKPDMJ, the concatenated sentence representation is used as the input to the next fully connected layer. The fully connected layer is composed of relu activation unit followed by a dropout (20 \%) and batch normalization. For all models, a final softmax layer is used to predict the output. The training of SKPDMJ model also followed the same pattern except the concatenation of SKP and DMJ features which is used as the input. The models are trained using `RMSProp' optimizer using a categorical cross-entropy loss function.  The number of fully connected layers and the learning rate were used as hyper-parameter. The learning rate we tried were in range $10^{-5}$ to $10^{-1}$. The fully-connected layer size we tried varied from $1$ to $3$. Once we find the best value for these hyper parameters by initial experiments, they remain unchanged during training and testing the performance of the model for all four events. For all models we find that a single fully connected layer performs better than multi-layered fully connected networks, so we use single layer network for all the results discussed next. 


\subsection{Results and Discussion}
We summarize the performance of the models in Tab. \ref{tbl:classification_results_for_all} in which we show the f1 score (micro) for all models for each dataset. As we can observe, if we consider the mean values across events, the replies-based models perform better. The performance is better not just when compared with quotes but also when compared with combined quotes and replies data. In fact, in all but one case, the model trained on combined data performs worse than both the replies based model and quotes based model. This confirms our earlier suspicion that people use quotes and replies in different ways on Twitter, and it is better to train separate models for inferring stance in quotes and replies.

If we compare the input features (Glove, SKP, DMJ, SKPDMJ), we can observe that most models are only slightly better than the majority (class) based model, which means that this problem is very challenging. The SVM model that used TF-IDF text features is the simplest yet performs as good as the deep learning models. Only on the combined data, the SVM is .01 worse than the Glove based model. This is not completely unexpected, as we know that most deep learning models require a lot of data to train, and in our case, we barely have a few thousand examples. What is more interesting is that even among the deep learning models, the Glove features based model that is the simplest to train, performs better than all other feature-based models. This is also unexpected given that earlier work, e.g., \cite{kumar2019tree}, has indicated the benefit of using sentence vectors (SKP, DMJ and SKPDMJ) in comparison to word vectors based models (GLove). This phenomenon could partially be because of the difference in the models used in the earlier work. 

\begin{figure}[htb!]
    \centering
    \includegraphics[width=0.49\textwidth]{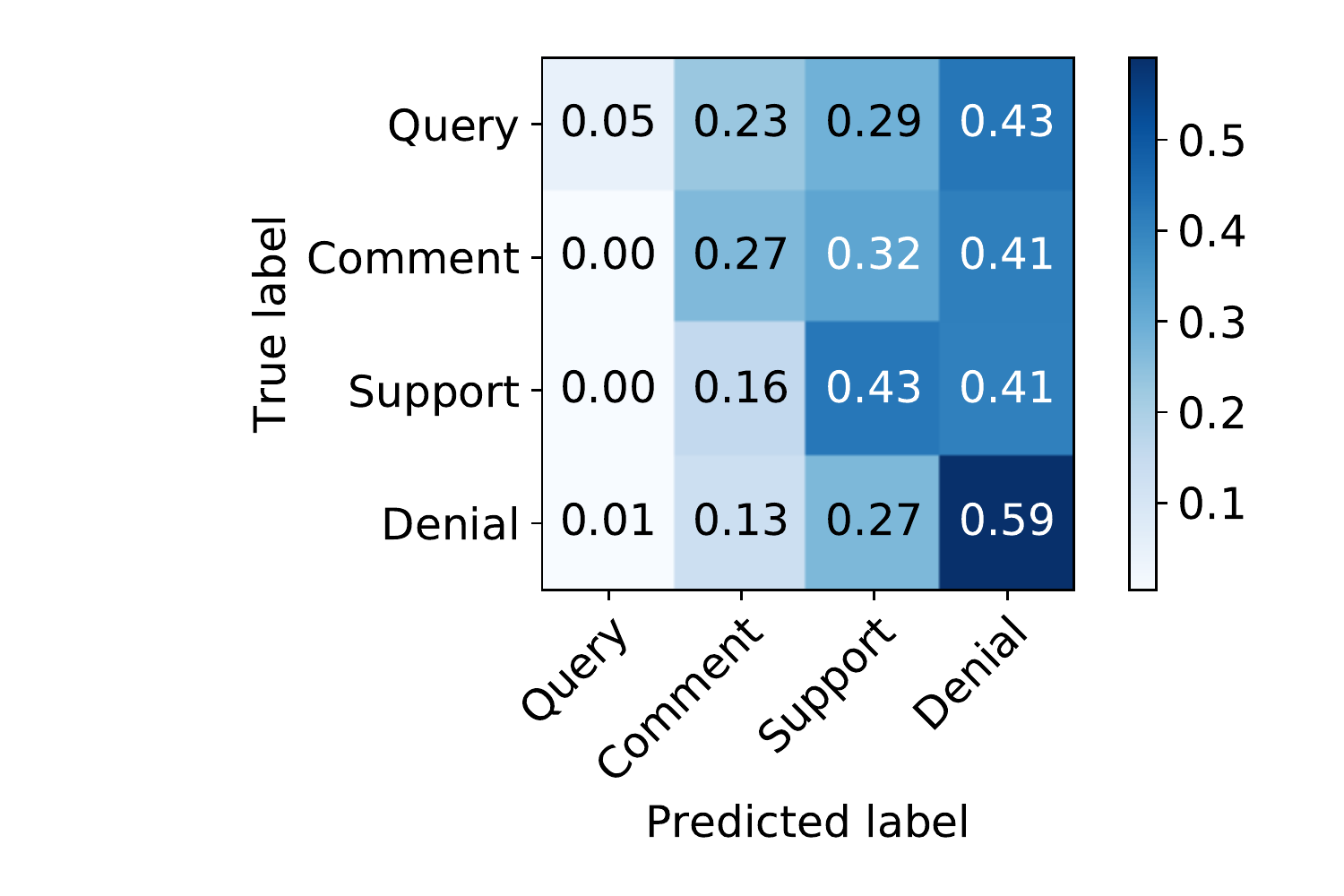}
    \caption{Confusion Matrix for Glove feature based deep-learning model for combined quotes and replies data.}
    \label{fig:confusion_matrix}
\end{figure}

If we consider the confusion matrix as shown in Fig. \ref{fig:confusion_matrix}, we can  observe that the `Denial' class is the best performing class followed by `support' class. This is aligned with the overall objective of this research to improve the denial class performance. In future work, we would like to combine the dataset prepared in earlier research \cite{zubiaga2015crowdsourcing} where `comment' is the majority class and and this new dataset that has more `Denial' and `Support' labels. 


%

\section{Conclusion and Future Work}
In this research, we created a new dataset that has stance labels for replies (and quotes) on Twitter posts on three controversial issues and on additional examples which do not belong to any specific topic. To overcome the limitations of prior research, we developed a collection methodology that is skewed toward non-neutral responses, and therefore has a more balanced class distribution as compared with prior datasets that have `Comment' as the majority class. We find that, when applied to contentious events, our methodology is effective at recovering contentious conversations and more non-neutral threads. Finally, our dataset also separates quotes and replies and is the first dataset to have stance labels for quotes. We envision that this dataset will allow other researchers to train and test models to automatically learn the stance taken by social-media users while replying to (or quoting) posts on social media.  

We also experimented with few machine learning models and evaluated their performance. We find that learning stance in conversations is still a challenging problem. Yet stance mining is important as  conversations are the only way to infer negative links between users of many platforms, and therefore inferring stance in conversations could be very valuable. We expect that our new dataset will allow the development of better stance learning models and enable a better understanding of community polarization and the detection of potential rumors.





\bibliographystyle{aaai}
\bibliography{main}

\onecolumn
\section{Appendix}\label{sec:appendix}

\begin{figure*}[ht!]
    \centering
    \frame{\includegraphics[width=0.9\textwidth]{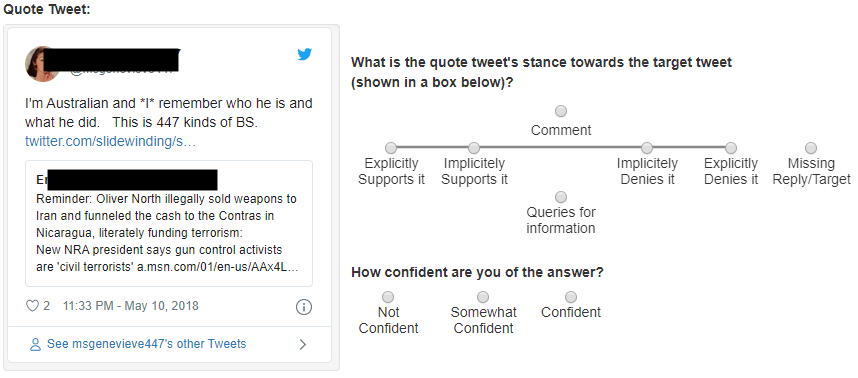}}
     \caption{Snapshot of the webpage developed for annotating quotes. Annotators are required to provide the stance in the quote and their confidence in the provided label. }
      \label{fig:annex_pages_quote}
\end{figure*}


 

\end{document}